\documentclass[11pt,a4paper]{article}
\usepackage{authblk}
\usepackage[hyperref]{emnlp2020}
\usepackage{times}
\usepackage{latexsym}

\usepackage{microtype}
\usepackage{bm}
\usepackage{graphicx}
\usepackage{pgfplots}
\pgfplotsset{compat=1.16}
\usepackage{tikz}
\usepackage{enumitem}
\setenumerate[1]{itemsep=0pt,partopsep=0pt,parsep=\parskip,topsep=2pt}
\setitemize[1]{itemsep=0pt,partopsep=0pt,parsep=\parskip,topsep=2pt}
\setdescription{itemsep=0pt,partopsep=0pt,parsep=\parskip,topsep=2pt}
\usepackage{cleveref}
\crefname{section}{§}{§§}
\aclfinalcopy 

\title{Dynamically Updating Event Representations for Temporal Relation Classification with Multi-category Learning}

\author[1]{Fei Cheng}
\author[2]{Masayuki Asahara}
\author[3]{Ichiro Kobayashi}
\author[1]{Sadao Kurohashi}
\affil[1]{Graduate School of Informatics, Kyoto University}
\affil[2]{National Institute for Japanese Language and Linguistics}
\affil[3]{Ochanomizu University}
\affil[ ]{\textit {\{feicheng,kuro\}@i.kyoto-u.ac.jp, masayu-a@ninjal.ac.jp, koba@is.ocha.ac.jp}}

\date{}

\begin{document}
\maketitle
\begin{abstract}
Temporal relation classification is a pair-wise task for identifying the relation of a temporal link (TLINK) between two mentions, i.e. event, time and document creation time (DCT). It leads to two crucial limits: 1) Two TLINKs involving a common mention do not share information. 2) Existing models with independent classifiers for each TLINK category (E2E, E2T and E2D)~\footnote{Time-to-Time (T2T) is not included in this paper, as we focus on event centric representations.}
hinder from using the whole data. This paper presents an event centric model that allows to manage dynamic event representations across multiple TLINKs. Our model deals with three TLINK categories with multi-task learning to leverage the full size of data. The experimental results show that our proposal outperforms state-of-the-art models and two transfer learning baselines on both the English and Japanese data.

\end{abstract}

\section{Introduction}
\label{sec:intro}

Reasoning over temporal relations relevant to an event mentioned in the document can help us understand when the event begins, how long it lasts, how frequent it is, and etc. Starting with the TimeBank~\citep{pustejovsky2003timebank} corpus, a series of temporal competitions (TempEval-1,2,3)~\citep{verhagen2009tempeval,verhagen2010semeval,uzzaman2012tempeval} are attracting growing research efforts.

Temporal relation classification (TRC) is the task to predict a temporal relation (\textit{after}, \textit{before}, \textit{includes}, etc.) of a TLINK from a source mention to a target mention. Less effort has been paid to explore the sharing information across `local' pairs and TLINK categories. In recent years, a variety of dense annotation schemas are proposed to overcome the `sparse' annotation in the original Timebank. A typical one is the Timebank-Dense (TD) corpus~\cite{chambers2014dense}, which performs a compulsory dense annotation with the complete graph of TLINKs for the mentions located in two neighbouring sentences. Such dense annotation increases the chance of pairs sharing common events and demands of managing `global' event representations across pairs among TLINK categories.

However, globally managing event representations of a whole document takes an extremely heavy load for the dense corpora. Timebank-Dense contains around 10,000 TLINKs in only 36 documents and is 7 times denser than the original Timebank. Thus, we propose a simplified scenario called Source Event Centric TLINK (SECT) chain. For each event $e_i$ in a document, we group all TLINKs containing the common source event $e_i$ into the $e_i$ centric TLINK chain and align them with the chronological order of the target mentions appearing in the document. We assume that our system is capable of learning dynamic representations of the centric event $e_i$ along the SECT chain via a `global' recurrent neural network (RNN).

\begin{quote}
    $DCT$: 1998-02-27 \\
    \textit{An intense \textbf{manhunt} (\bm{$e_1$}) conducted by the FBI and the bureau of alcohol, tobacco and firearms \textbf{continues} (\bm{$e_2$}) for Rudolph in the wilderness of western north Carolina.}
    \textit{And \textbf{this week} (\bm{$t_1$}), FBI director Louie Freeh assigned more agents to the \textbf{search} (\bm{$e_3$}).}
\end{quote}

We demonstrate our proposal with the above adjacent-sentence excerpt in Timebank-Dense. `$(e_s, e_t)$' denotes a directed TLINK from the source $e_s$ to target $e_t$ in this paper. Considering the `\textit{\textbf{manhunt}} ($e_1$)' centric chain: $\{(e_1, DCT), (e_1, e_2), (e_1, t_1), (e_1, e_3)\}$\footnote{As DCT is not explicitly mentioned in documents, we always place $(e_i, DCT)$ on the top of a SECT chain}, `\textit{\textbf{manhunt}}' holds a `\textit{includes}' relation to `\textit{\textbf{continues}}'. We assume that dynamically updating the representation of `\textit{\textbf{manhunt}}' in the early step `$(e_1, e_2)$' will benefit the prediction for the later step $(e_1, e_3)$ to `\textit{\textbf{search}}'. `\textit{\textbf{manhunt}}' is supposed to hold the same `\textit{includes}' relation to  `\textit{\textbf{search}}', as the search should be included in the continuing manhunt.

Our model further exploits a multi-task learning framework to leverage all three categories of TLINKs in the SECT chain scope. A common BERT~\citep{devlin-etal-2019-bert} encoder layer is applied to retrieve token embeddings. The global RNN layer manages the dynamic event and TLINK presentations in the chain.  Finally, our system feeds the TLINK representations into their corresponding category-specific (E2D, E2T and E2E) classifiers to calculate a combined loss.

The contribution of this work is listed as follows:
1) We present a novel source event centric model to dynamically manage event representations across TLINKs. 
2) Our model exploits a multi-task learning framework with two common layers trained by a combined category-specific loss to overcome the data isolation among TLINK categories.
The experimental results suggest the effectiveness of our proposal on two datasets.
All the codes of our model and two baselines is released.~\footnote{\url{https://github.com/racerandom/NeuralTime}}

\section{Related Work}
\label{sec:rw}

\subsection{Temporal Relation Classification}
\label{sec:rw:trc}

Most existing temporal relation classification approaches focus on extracting various features from the textual sentence in the local pair-wise setting. Inspired by the success of neural networks in various NLP tasks, \citet{cheng-miyao:2017:Short,meng-rumshisky-romanov:2017:EMNLP2017,vashishtha-etal-2019-fine,han-etal-2019-joint,han-etal-2019-deep} propose a series of neural networks to achieve accuracy with less feature engineering. However, these neural models still drop in the pair-wise setting.

\citet{meng-rumshisky-2018-context} propose a global context layer (GCL) to store/read the solved TLINK history upon a pre-trained pair-wise classifier. However, they find slow converge when training the GCL and pair-wise classifier simultaneously. Minor improvement is observed compared to their pair-wise classifier. Our model is distinguished from their work in three focuses: 1) We constrains the model in a reasonable scope, i.e. SECT chain. 2) We manages dynamic event representations, while their model stores/reads pair history 3) Our model integrates category-specific classifiers by multi-task learning, while they use the categories as the features in one single classifier.

\subsection{Multi-task Transfer Learning}
\label{sec:rw:tl}

For the past three years, several successful transfer learning models (ELMO, GPT and BERT)~\citep{peters-etal-2018-deep,radford2018improving,devlin-etal-2019-bert} have been proposed, which significantly improved the state-of-the-art on a wide range of NLP tasks. \cite{liu-etal-2019-multi} propose a single-task batch multi-task learning approach over a common BERT to leverage a large mount of cross-task data in the fine-tuning stage.

In this work, our model deals with various categories of TLINKs (E2E, E2T and E2D)  in a batch of SECT chains to calculate the combined loss with the category-specific classifiers. 

\subsection{Non-English Temporal Corpora}

Less attention has been paid for non-English temporal corpora. Until 2014, \citeauthor{asahara2014bccwj} starts the first corpus-based study BCCWJ-Timebank (BT) on Japanese temporal information annotation. We explore the feasibility of our model on this Japanese dataset. 

\begin{figure}[t]
\center
\includegraphics[width=1.0\linewidth]{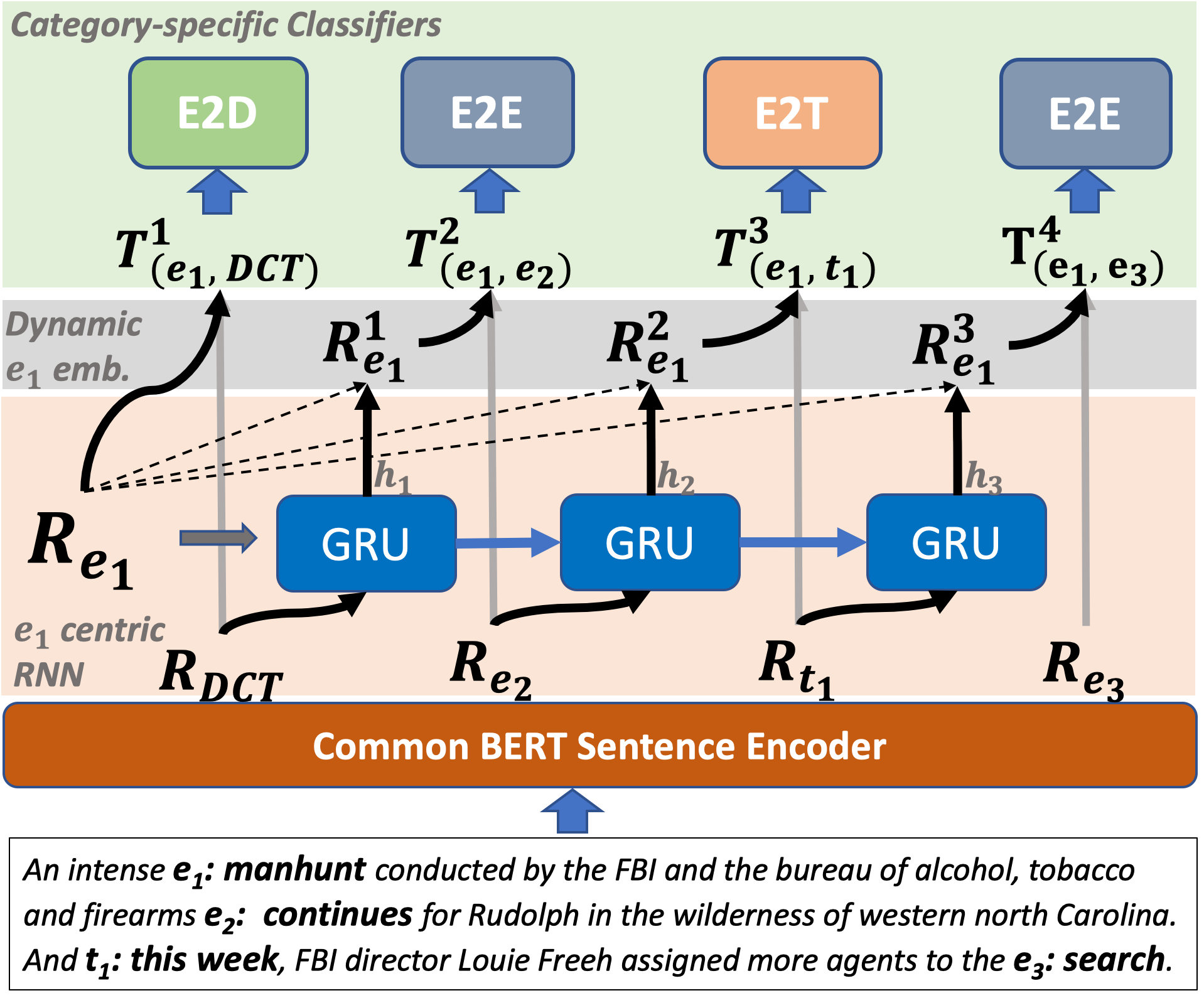}
\caption{\label{fig:system} The overview of the proposed model. }
\end{figure}

\begin{figure*}[t]
   \centering

\begin{tikzpicture}[yscale=0.55, xscale=0.65]
\begin{axis}[
    xmin=1, xmax=20,
    ymin=30, ymax=70,
    xlabel=training epochs (Timebank-Dense), ylabel=Dev F1, 
    xtick={2,4,6,8,10,12,14,16,18,20},
    ytick={30, 40,50,60,60},
    legend pos=north west,
    ymajorgrids=true,
    grid style=dashed,
]

\addplot[
    color=black,
    mark=square,
    ]
    coordinates {
    (1,40.61)(2,48.10)(3,53.31)(4,52.55)(5,50.27)(6, 49.73)(7,48.12)(8,48.67)(9,52.03)(10,50.73)(11,52.14)(12, 50.29)(13,51.38)(14,48.10)(15,49.95)(16,48.10)(17,49.19)(18,49.73)(19,49.84)(20,49.40)
    };
\addplot[
    color=blue,
    mark=square,
    ]
    coordinates {
    (1,35.79)(2,37.55)(3,38.44)(4,40.72)(5,42.67)(6, 43.97)(7,44.41)(8,44.73)(9,44.73)(10,46.04)(11,47.71)(12,46.73)(13,47.28)(14,46.95)(15,47.99)(16,47.12)(17,48.53)(18,47.01)(19,47.12)(20,46.15)
    };
\addplot[
    color=red,
    mark=square,
    ]
    coordinates {
    (1,40.61)(2,48.10)(3,53.31)(4,54.03)(5,54.72)(6, 54.51)(7,54.94)(8,54.29)(9,55.16)(10,55.92)(11,56.03)(12,56.68)(13,56.79)(14,56.57)(15,56.46)(16,57.44)(17,57.11)(18,57.00)(19,56.35)(20,56.79)
    };
\legend{no freeze, freeze, freeze after $k$ epochs}
\end{axis}
\end{tikzpicture}
~~~~~~~~~%
\begin{tikzpicture}[yscale=0.55, xscale=0.65]
\begin{axis}[
    xmin=1, xmax=20,
    ymin=40, ymax=90,
    xlabel=training epochs (BCCWJ-Timebank),ylabel=Dev F1, 
    xtick={2,4,6,8,10,12,14,16, 18,20},
    ytick={40,50,60, 70, 80},
    legend pos=north west,
    ymajorgrids=true,
    grid style=dashed,
]

\addplot[
    color=black,
    mark=square,
    ]
    coordinates {
    (1,56.59)(2,62.97)(3,67.18)(4,67.86)(5,67.83)(6, 67.43)(7,67.71)(8,65.53)(9,64.90)(10,66.19)(11,67.54)(12,66.95)(13,67.51)(14,67.31)(15,68.35)(16,66.48)(17,67.93)(18,65.73)(19,65.85)(20,64.11)
    };
\addplot[
    color=blue,
    mark=square,
    ]
    coordinates {
    (1,46.52)(2,53.45)(3,58.60)(4,60.24)(5,62.14)(6, 64.38)(7,65.72)(8,66.03)(9,66.27)(10,65.42)(11,65.90)(12,65.46)(13,65.13)(14,65.77)(15,65.98)(16,66.50)(17,65.49)(18,65.37)(19,65.59)(20,65.47)
    };
\addplot[
    color=red,
    mark=square,
    ]
    coordinates {
    (1,56.59)(2,62.97)(3,67.18)(4,67.65)(5,68.60)(6, 69.14)(7,69.54)(8,69.66)(9,69.61)(10,69.80)(11,69.86)(12,69.56)(13,69.29)(14,69.49)(15,69.59)(16,69.64)(17,70.56)(18,68.69)(19,70.06)(20,69.89)
    };
\legend{no freeze, freeze, freeze after $k$ epochs}
\end{axis}
\end{tikzpicture}
\caption{\label{fig:ats}Dev performance (micro-F1) of three training strategies on two datasets.}
\end{figure*}
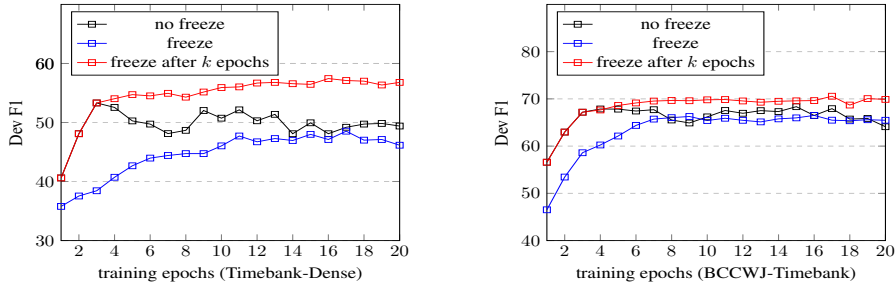

\section{Overview of Proposed Model}
\label{sec:ger}

Figure~\ref{fig:system} demonstrates the overview of our Source Event Centric (SEC) model with the previous $e_1$ centric chain example $\{(e_1, DCT), (e_1, e_2), (e_1, t_1), (e_1, e_3)\}$ in \cref{sec:intro}.

\subsection{BERT Sentence Encoder}

  We apply a pre-trained BERT for retrieving token embeddings of input sentences. 
For a multiple-token mention, we treat the element-wise sum of token embeddings as the mention embedding.

\subsection{Source Event Centric RNN}

After the BERT layer processing, the system collects all the  mention embeddings appearing in the chain: $\{R_{e_1}, R_{DCT}, R_{e_2}, R_{t_1}, R_{e_3}\}$\footnote{As DCT is not explicitly mentioned in documents, we set $R_{DCT}$ as a trainable embedding.}. 

Our model assigns a `global' two-layer gated recurrent unit (GRU) model with the left-to-right direction to simulate the chronological order of the SETC chain for updating the centric $e_1$ embeddings. 
The original $e_1$ embedding $R_{e_1}$ is sent into the GRU as the initial hidden. At $i$-th TLINK step, the system inputs the target mention embedding to update the $i$-th $e_1$ embedding $R_{e_1}^{i}$ for generating the $\{i+1\}$-th step TLINK embedding $T^{i+1}$. As shown in Figure~\ref{fig:system}, the $3$-rd TLINK embedding $T_{(e_1, t_1)}^{3}$ is the concatenation of the $2$-nd step $R_{e_1}^{2}$ and target embedding $R_{t_1}$ as the follows:

\noindent
\begin{equation}
R_{e_1}^{2} = max(R_{e_1}, GRU(R_{e_2}, h_{1}))
\end{equation}

\noindent
\begin{equation}
T_{(e_1, t_1)}^{3} = [R_{e_1}^{2};R_{t_1}]
\end{equation}

The element-wise $max$ is desiged to set the initial $R_{e_1}$ as an anchor to avoid the quality dropping of new hiddens after long sequential updating.


\subsection{Multi-category Learning}

After obtaining all the TLINK embeddings $\{T_{(e_1, DCT)}^{1}, T_{(e_1, e_2)}^{2}, T_{(e_1, t_1)}^{3}, T_{(e_1,e_3)}^{4}\}$ in the SECT chain via the previous two common layers, the system feeds them into the corresponding category-specific classifiers. Each classifier is built with one linear full-connected layer and Softmax layer. The system calculates the combined loss as the follows to perform multi-category learning.

\noindent
\begin{equation}
    L = L_{E2E} + L_{E2T} + L_{E2D}
\end{equation}


\section{Experiments and Results}
\label{sec:exp}

We conduct the experiments of applying the SEC model on both the English TD and Japanese BT corpora. Juman++~\cite{tolmachev-etal-2018-juman}\footnote{\url{https://github.com/ku-nlp/jumanpp}} is adopted to do morphological analysis for Japanese text. TD annotation adopts a 6-relation set (\textit{after}, \textit{before}, \textit{simultaneous}, \textit{includes}, \textit{is\_included} and \textit{vague}). We follow the `train/dev/test` data split\footnote{ \url{www.usna.edu/Users/cs/nchamber/caevo}} of the previous work. For BT, we follow a merged 6-relation set as~\cite{Yoshikawa2014}. We perform the document-level 5-fold cross-validation. In each split, we randomly select 15\% documents as the dev set from the training set. The TLINKs statistics of the two corpora are listed in Table~\ref{tab:stats}.

\begin{table}[t]
\centering
\begin{tabular}{l|r@{\hskip5pt}r@{\hskip5pt}r@{\hskip6pt}r@{\hskip6pt}|r}
\hline
\textbf{ Corpus} & \textbf{E2D} & \textbf{E2T} & \textbf{E2E} & \textbf{MAT} & \textbf{SECT}\\
\hline
English & 1,494 & 2,001 & 6,088 & - & 5.5 \\
Japanese & 2,873 & 1,469 & 1,862 & 776 & 2.4 \\
\hline
\end{tabular}
\caption{\label{tab:stats} Number of TLINKs in the English and Japanese corpora. `SECT` denotes the average TLINK number per SECT chain. `MAT' is defined in \cref{sec:exp:ja}}
\end{table}

We adopt the English and Japanese pre-trained `base' BERT\footnote{\url{github.com/huggingface/transformers}} and empirically set RNN hidden size equal to BERT hidden, 4 SECT chains per batch, 20 epochs, and AdamW (lr=5e-5). The other hyper-parameters are selected based on the dev micro-F1. All the results are 5-run average.


For the lack of comparable transfer learning approaches, we build two BERT baselines as follows (fine-tuning 5 epochs, batch size is 16):
\begin{itemize}
    \itemsep0em
    \item {\textbf{Local-BERT}: The concatenation of two mentions as TLINK embeddings are fed into the independent category-specific classifier.}
    \item {\textbf{Multi-BERT}: The multi-category setting as \cite{liu-etal-2019-multi} of Local-BERT.  Each time the system pops out a single-category batch, encodes it via the common BERT, and feed it to the category-specific classifier. }
\end{itemize}

`Local-BERT' and `Multi-BERT' serve as the baselines in the ablation test for the proposed `SEC' model. `Local-BERT' is the `SEC' model removing both global RNN and multi-category learning. `Multi-BERT' is viewed as the `SEC' model removing global RNN.

\subsection{Asynchronous Training Strategy}
Fine-tuning BERT is difficultly performed with training SEC RNN simultaneously. The standard fine-tuning only requires 3 to 5 epochs, which indicates the pre-trained model tends to quickly overfit. However, the SEC RNN is randomly initialized and requires more training epochs. 
\begin{itemize}
    \item {\textbf{no freeze} of BERT sentence encoder}
    \item {\textbf{freeze} of BERT sentence encoder}
    \item {\textbf{freeze} \textbf{after $\bm{k}$ epochs}}
\end{itemize}

Figure~\ref{fig:ats} shows the validation micro F1 of all TLINKs against the training epochs of the above asynchronous training strategies.  \textbf{no freeze} shows the evidence of our concern that the curve undulate after the initial 3 epochs. \textbf{freeze} performs a stable learning phase with the lowest initialization. \textbf{freeze after $\bm{k}$ epochs} achieves the balance of the stability and high F1. Therefore, we perform the third strategy for all the following experiments. The number $k$ is selected from $\{3, 4, 5\}$ based on the validation scores.

\subsection{Main Timebank-Dense Results}
\label{sec:exp:main}

Table~\ref{tab:main} shows the experimental results on the English TD corpus. `CATENA'~\cite{mirza2016} is the feature-based model combined with dense word embeddings. `SDP-RNN'~\cite{cheng-miyao:2017:Short} is the dependency tree enhanced RNN model.`GCL'~\cite{meng-rumshisky-2018-context} is the global context layer model introduced in \cref{sec:rw:trc}. `Fine-grained TRC'~\citet{vashishtha-etal-2019-fine} is the ELMO based fine-grained TRC model with only the E2E results reported.

It's not surprising that the proposed model substantially outperforms state-of-the-art systems, as the existing SOTA didn't exploit BERT yet. Therefore, we offer the ablation test with ‘Local-BERT’(w/o multi-categories learning and global SEC RNN) and ‘Multi-BERT’ (w/o global SEC RNN) to investigate the benefits of our two contributions. The `SEC' model obtains +3.2, +6.8, +5.2 F1 improvements compared to `Local-BERT', which suggests the effectiveness of two main proposal. The `SEC' model further outperforms ‘Multi-BERT’ by 3.6 gain of the majority category E2E, 1.0 gain of E2T and 0.7 gain of E2D, which indicates the impact of the global SEC RNN. 

A main finding is that E2E obtains higher gains from `global' contexts, compare to E2T and E2D. It matches the intuition that events are more globally contextualized and time expressions are usually more self-represented (e.g. normalized time values). E2D mainly requires contextual information from the single sentences by the BERT encoder. E2T takes less advantage of BERT, while multi-category training with E2E, E2D can significantly improves its performance. 

\begin{table}[t]
\centering
\begin{tabular}{l@{\hskip6pt}rrrr}
\hline
\textbf{Models} & \textbf{E2D} & \textbf{E2T} & \textbf{E2E}\\
\hline
Majority Vote & 32.3 & 40.6 & 47.7 \\
\hline
\multicolumn{4}{l}{\textit{local Models}} \\
CATENA (2016) & 53.4 & 46.8 & 51.9 \\
SDP-RNN (2017) & 54.6 & 47.1 & 52.9 \\
Fine-grained TRC (2019) & - & - & 56.6 \\
Local-BERT & 62.7 & 49.4 & 59.8 \\
\hline
\multicolumn{4}{l}{\textit{local + multi-category Models}} \\
Multi-BERT & 65.2 & 54.8 & 61.4  \\
\hline
\multicolumn{4}{l}{\textit{global + multi-category Models}} \\
GCL (2018) & 48.9 & 48.7 & 57.0  \\
\textbf{SEC (proposed)} & \textbf{65.9} & \textbf{55.8} & \textbf{65.0} \\
\hline
\end{tabular}
\caption{\label{tab:main} Temporal relation classification results (micro F1) on the English Timebank-Dense. }
\end{table}

\subsection{Results on Non-English Data}
\label{sec:exp:ja}

\begin{table}[t]
\centering
\begin{tabular}{l@{\hskip6pt}rrrr}
\hline
\textbf{Models} & \textbf{E2D} & \textbf{E2T} & \textbf{E2E} & \textbf{MAT}\\
\hline
Majority Vote & 68.3 & 50.4 & 43.2 & 39.3 \\
\hline
\multicolumn{4}{l}{\textit{local Models}} \\
Yoshikawa (2014) & 75.6 & 55.7 & 59.9 & 50.0 \\
Local-BERT & 80.7 & 58.9 & 61.2 & 54.1 \\
\hline
\multicolumn{4}{l}{\textit{local + multi-category Models}} \\
Multi-BERT & 81.4 & \textbf{61.0} & 63.3 & 61.6 \\
\hline
\multicolumn{4}{l}{\textit{global + multi-category  Models}} \\
\textbf{SEC (proposed)} & \textbf{81.6} & 60.7 & \textbf{64.5} & \textbf{64.6} \\
\hline
\end{tabular}
\caption{\label{jaresult} Temporal relation classification results (micro F1) on the Japanese BCCWJ-Timebank.}
\end{table}

Table~\ref{jaresult} shows the results in the Japanese corpus. Different from 
the TD annotation schema, BT specifies two E2E categories for fitting the Japanese language:  1) E2E: between two consecutive events, 2) MAT: between two consecutive matrix verb events.

The state-of-the-art system on BT is the feature-based approach~\cite{Yoshikawa2014}. The comparisons are similar to the English data. Our `SEC` obtains the substantial improvements compared to their work and two BERT baselines. An interesting observation is that MAT TLINKs are usually inter-sentence located at the end of SECT chains, as Japanese is a `SOV' language. The results indicate that long distance MAT suffers from the low-quality representations in the `local' setting and benefits from `global' representation more.

\section{Conclusion}
\label{sec:con}
This paper presents a novel transfer learning based model to boost the performance of temporal information extraction task especially for densely annotated dataset.  Our model can dynamically update event representations across multiple TLINKs in a Source Event Centric chain scope. Our model exploits a multi-category learning framework to leverage the total data of three TLINK categories. The empirical results show that our proposal outperforms the state-of-the-art systems and the ablation tests suggest the effectiveness of two main proposals. The Non-English experiments support the feasibility of our system on the Japanese data.

\bibliography{anthology,emnlp2020}

\begin{thebibliography}{19}
\expandafter\ifx\csname natexlab\endcsname\relax\def\natexlab#1{#1}\fi

\bibitem[{Asahara et~al.(2014)Asahara, Kato, Konishi, Imada, and
  Maekawa}]{asahara2014bccwj}
Masayuki Asahara, Sachi Kato, Hikari Konishi, Mizuho Imada, and Kikuo Maekawa.
  2014.
\newblock Bccwj-timebank: Temporal and event information annotation on japanese
  text.
\newblock In \emph{International Journal of Computational Linguistics \&
  Chinese Language Processing, Volume 19, Number 3, September 2014}.

\bibitem[{Chambers et~al.(2014)Chambers, Cassidy, McDowell, and
  Bethard}]{chambers2014dense}
Nathanael Chambers, Taylor Cassidy, Bill McDowell, and Steven Bethard. 2014.
\newblock \href {http://aclweb.org/anthology/Q/Q14/Q14-1022.pdf} {Dense event
  ordering with a multi-pass architecture}.
\newblock \emph{Transactions of the Association for Computational Linguistics},
  2:273--284.

\bibitem[{Cheng and Miyao(2017)}]{cheng-miyao:2017:Short}
Fei Cheng and Yusuke Miyao. 2017.
\newblock \href {http://aclweb.org/anthology/P17-2001} {Classifying temporal
  relations by bidirectional lstm over dependency paths}.
\newblock In \emph{Proceedings of the 55th Annual Meeting of the Association
  for Computational Linguistics (Volume 2: Short Papers)}, pages 1--6,
  Vancouver, Canada. Association for Computational Linguistics.

\bibitem[{Devlin et~al.(2019)Devlin, Chang, Lee, and
  Toutanova}]{devlin-etal-2019-bert}
Jacob Devlin, Ming-Wei Chang, Kenton Lee, and Kristina Toutanova. 2019.
\newblock \href {https://doi.org/10.18653/v1/N19-1423} {{BERT}: Pre-training of
  deep bidirectional transformers for language understanding}.
\newblock In \emph{Proceedings of the 2019 Conference of the North {A}merican
  Chapter of the Association for Computational Linguistics: Human Language
  Technologies, Volume 1 (Long and Short Papers)}, pages 4171--4186,
  Minneapolis, Minnesota. Association for Computational Linguistics.

\bibitem[{Han et~al.(2019{\natexlab{a}})Han, Hsu, Yang, Galstyan, Weischedel,
  and Peng}]{han-etal-2019-deep}
Rujun Han, I-Hung Hsu, Mu~Yang, Aram Galstyan, Ralph Weischedel, and Nanyun
  Peng. 2019{\natexlab{a}}.
\newblock \href {https://doi.org/10.18653/v1/K19-1062} {Deep structured neural
  network for event temporal relation extraction}.
\newblock In \emph{Proceedings of the 23rd Conference on Computational Natural
  Language Learning (CoNLL)}, pages 666--106, Hong Kong, China. Association for
  Computational Linguistics.

\bibitem[{Han et~al.(2019{\natexlab{b}})Han, Ning, and
  Peng}]{han-etal-2019-joint}
Rujun Han, Qiang Ning, and Nanyun Peng. 2019{\natexlab{b}}.
\newblock \href {https://doi.org/10.18653/v1/D19-1041} {Joint event and
  temporal relation extraction with shared representations and structured
  prediction}.
\newblock In \emph{Proceedings of the 2019 Conference on Empirical Methods in
  Natural Language Processing and the 9th International Joint Conference on
  Natural Language Processing (EMNLP-IJCNLP)}, pages 434--444, Hong Kong,
  China. Association for Computational Linguistics.

\bibitem[{Liu et~al.(2019)Liu, He, Chen, and Gao}]{liu-etal-2019-multi}
Xiaodong Liu, Pengcheng He, Weizhu Chen, and Jianfeng Gao. 2019.
\newblock \href {https://doi.org/10.18653/v1/P19-1441} {Multi-task deep neural
  networks for natural language understanding}.
\newblock In \emph{Proceedings of the 57th Annual Meeting of the Association
  for Computational Linguistics}, pages 4487--4496, Florence, Italy.
  Association for Computational Linguistics.

\bibitem[{Meng and Rumshisky(2018)}]{meng-rumshisky-2018-context}
Yuanliang Meng and Anna Rumshisky. 2018.
\newblock \href {https://doi.org/10.18653/v1/P18-1049} {Context-aware neural
  model for temporal information extraction}.
\newblock In \emph{Proceedings of the 56th Annual Meeting of the Association
  for Computational Linguistics (Volume 1: Long Papers)}, pages 527--536,
  Melbourne, Australia. Association for Computational Linguistics.

\bibitem[{Meng et~al.(2017)Meng, Rumshisky, and
  Romanov}]{meng-rumshisky-romanov:2017:EMNLP2017}
Yuanliang Meng, Anna Rumshisky, and Alexey Romanov. 2017.
\newblock \href {https://www.aclweb.org/anthology/D17-1092} {Temporal
  information extraction for question answering using syntactic dependencies in
  an lstm-based architecture}.
\newblock In \emph{Proceedings of the 2017 Conference on Empirical Methods in
  Natural Language Processing}, pages 887--896, Copenhagen, Denmark.
  Association for Computational Linguistics.

\bibitem[{Mirza and Tonelli(2016)}]{mirza2016}
Paramita Mirza and Sara Tonelli. 2016.
\newblock \href {http://aclweb.org/anthology/C16-1265} {On the contribution of
  word embeddings to temporal relation classification}.
\newblock In \emph{Proceedings of COLING 2016, the 26th International
  Conference on Computational Linguistics: Technical Papers}, pages 2818--2828,
  Osaka, Japan. The COLING 2016 Organizing Committee.

\bibitem[{Peters et~al.(2018)Peters, Neumann, Iyyer, Gardner, Clark, Lee, and
  Zettlemoyer}]{peters-etal-2018-deep}
Matthew Peters, Mark Neumann, Mohit Iyyer, Matt Gardner, Christopher Clark,
  Kenton Lee, and Luke Zettlemoyer. 2018.
\newblock \href {https://doi.org/10.18653/v1/N18-1202} {Deep contextualized
  word representations}.
\newblock In \emph{Proceedings of the 2018 Conference of the North {A}merican
  Chapter of the Association for Computational Linguistics: Human Language
  Technologies, Volume 1 (Long Papers)}, pages 2227--2237, New Orleans,
  Louisiana. Association for Computational Linguistics.

\bibitem[{Pustejovsky et~al.(2003)Pustejovsky, Hanks, Sauri, See, Gaizauskas,
  Setzer, Radev, Sundheim, Day, Ferro et~al.}]{pustejovsky2003timebank}
James Pustejovsky, Patrick Hanks, Roser Sauri, Andrew See, Robert Gaizauskas,
  Andrea Setzer, Dragomir Radev, Beth Sundheim, David Day, Lisa Ferro, et~al.
  2003.
\newblock \href {https://catalog.ldc.upenn.edu/LDC2006T08} {The timebank
  corpus}.
\newblock In \emph{Corpus linguistics}, volume 2003, page~40.

\bibitem[{Radford et~al.(2018)Radford, Narasimhan, Salimans, and
  Sutskever}]{radford2018improving}
Alec Radford, Karthik Narasimhan, Tim Salimans, and Ilya Sutskever. 2018.
\newblock Improving language understanding by generative pre-training.

\bibitem[{Tolmachev et~al.(2018)Tolmachev, Kawahara, and
  Kurohashi}]{tolmachev-etal-2018-juman}
Arseny Tolmachev, Daisuke Kawahara, and Sadao Kurohashi. 2018.
\newblock \href {https://doi.org/10.18653/v1/D18-2010} {{J}uman++: A
  morphological analysis toolkit for scriptio continua}.
\newblock In \emph{Proceedings of the 2018 Conference on Empirical Methods in
  Natural Language Processing: System Demonstrations}, pages 54--59, Brussels,
  Belgium. Association for Computational Linguistics.

\bibitem[{UzZaman et~al.(2012)UzZaman, Llorens, Allen, Derczynski, Verhagen,
  and Pustejovsky}]{uzzaman2012tempeval}
Naushad UzZaman, Hector Llorens, James Allen, Leon Derczynski, Marc Verhagen,
  and James Pustejovsky. 2012.
\newblock \href {http://aclweb.org/anthology/S/S13/S13-2001.pdf} {Tempeval-3:
  Evaluating events, time expressions, and temporal relations}.
\newblock \emph{arXiv preprint arXiv:1206.5333}.

\bibitem[{Vashishtha et~al.(2019)Vashishtha, Van~Durme, and
  White}]{vashishtha-etal-2019-fine}
Siddharth Vashishtha, Benjamin Van~Durme, and Aaron~Steven White. 2019.
\newblock \href {https://doi.org/10.18653/v1/P19-1280} {Fine-grained temporal
  relation extraction}.
\newblock In \emph{Proceedings of the 57th Annual Meeting of the Association
  for Computational Linguistics}, pages 2906--2919, Florence, Italy.
  Association for Computational Linguistics.

\bibitem[{Verhagen et~al.(2009)Verhagen, Gaizauskas, Schilder, Hepple,
  Moszkowicz, and Pustejovsky}]{verhagen2009tempeval}
Marc Verhagen, Robert Gaizauskas, Frank Schilder, Mark Hepple, Jessica
  Moszkowicz, and James Pustejovsky. 2009.
\newblock \href {https://doi.org/10.1007/s10579-009-9086-z} {The tempeval
  challenge: identifying temporal relations in text}.
\newblock \emph{Language Resources and Evaluation}, 43(2):161--179.

\bibitem[{Verhagen et~al.(2010)Verhagen, Sauri, Caselli, and
  Pustejovsky}]{verhagen2010semeval}
Marc Verhagen, Roser Sauri, Tommaso Caselli, and James Pustejovsky. 2010.
\newblock \href {http://www.aclweb.org/anthology/S10-1010} {Semeval-2010 task
  13: Tempeval-2}.
\newblock In \emph{Proceedings of the 5th international workshop on semantic
  evaluation}, pages 57--62. Association for Computational Linguistics.

\bibitem[{Yoshikawa et~al.(2014)Yoshikawa, Asahara, and Iida}]{Yoshikawa2014}
Katsumasa Yoshikawa, Masayuki Asahara, and Ryu Iida. 2014.
\newblock Estimating temporal order relation for bccwj-timebank.
\newblock In \emph{Proceedings of the Japanese Annual Conference on NLP}.
\newblock (in Japanese).

\end{thebibliography}
\bibliographystyle{acl_natbib}

\end{document}